\def\year{2015}
\documentclass[letterpaper]{article}
\usepackage{aaai}
\usepackage{times}
\usepackage{helvet}
\usepackage{courier}

\usepackage{graphicx}
\usepackage{floatrow}
\usepackage{natbib}
\usepackage{comment}
\begin{document}
%
\title{Predicting the top and bottom ranks of billboard songs using Machine Learning} \author{Vivek Datla \and Abhinav Vishnu \\
Pacific Northwest National Laboratory\\
vivek.datla@pnnl.gov \and abhinav.vishnu@pnnl.gov\\
Richland, WA 99354\\
}
\maketitle
\begin{abstract}
The music industry is a \$130 billion industry. Predicting whether a song catches the pulse of the audience impacts the industry. In this paper we analyze language inside the lyrics of the songs using several computational linguistic algorithms and predict whether a song would make to the top or bottom of the billboard rankings based on the language features. 

We trained and tested an SVM classifier with a radial kernel function on the linguistic features. Results indicate that we can classify whether a song belongs to top and bottom of the billboard charts with a precision of 0.76.
\end{abstract}

\section {Introduction}

German philosopher Friedrich Nietzche famously said \emph{``without music, life would be a mistake''}. In this digital age, we have access to a large collection of music composed at an amazing rate. iTunes music store alone offers 37 million songs, and has sold more than 25 billion songs worldwide.

Every society has its version of music and popularity of the songs, and sometimes they transcend the societies as well as continents. The 90\'s era of pop and rock music was dominated by artists such as Micheal Jackson, Sting, U2 and many others. The whole generation of 90\'s
youth can immediately identify \emph{``Beat it!"} a top song during that period.

What makes a song catchy? The lyrics of the songs contain words that arouse several emotions such as anger, and love, which tend to play an important role in humans liking the songs. The
liking of the songs does have not only a human emotion aspect but also has a direct economic impact on the \$130 billion music industry. 

The sales and evaluation of the songs directly impact the music companies and a computational model that predicts the popularity of a song is of great value for the music industry. Identifying the potential of a song earlier gives an edge for the companies to purchase the songs at a lower cost. Also, an artist usually composes the music for a song after the lyrics are written. For an organization investing in a music album, it is a great financial incentive to know whether the song would catch the pulse of the audience just based on the lyrics even before the music album is composed, as composing music requires considerable resources.

Since songs are composed of several complex components such as lyrics, instrumental music, vocal and visual renditions, the nature of a song itself is highly complex. Lyrics is the language component that ties up the vocal, music, and visual components. There needs to be harmony between the components to produce a song. Songs have the potential to lift our moods, make us shake a leg or move us to tears. They also help us relate to our experiences, by triggering several emotional responses.

There has been a lot of work on genre classification using machine learning. Researchers identify the category of the songs based on the emotions such as sad, happy and party. All the songs tend to have an emotional component, but we see very few songs that catch the people's pulse and become a hit. 

The research question addressed in the paper are as follows:

\begin{itemize}
  \item Can machine learning models be trained on lyrics for predicting the top and bottom ranked songs?
\end{itemize}


In the current paper, we look at language features that help predict whether a song belongs to a top or a bottom ranked category. To the best of our knowledge, this is the first study addressing this problem.

\section{Related Work}

Language is a strong indicator of stresses and mood of a person. Identifying these features has helped computational linguists as well as computer scientists to correlate the language features with several complex problems arising in tutoring systems \citep{rus2013,graesser2005}, affect recognition\citep{dmello2008},
sentiment mining ~\citep{Hu:2004}, opinion mining, and many others. 

~\cite{su2013} implemented a multimodal music emotion classification (MEC)  for classifying 14 kinds of emotions from music and song lyrics of western music genre. Their dataset consisted of \~3500 songs with emotions/mood such as sad, high, groovy, happy, lonely, sexy, energetic, romantic, angry, sleepy, nostalgic, funny, jazzy, and calm. They used AdaBoost with decision stumps for classification of the music and language features of the lyrics into their respective emotion categories. They have an accuracy of ~0.78 using language as well as surface features of the audio. The authors claim that the language features played a more important role compared to the music features in classification.

~\cite{laurier2008} also indicated that the language features outperformed audio features for music mood classification. They have
shown that language features extracted from the songs fit well with Russel's valence(negative-positive) and arousal(inactive-active) model
~\citep{russell1980}. Several cross-cultural studies show evidence for universal emotional cues in music and language across different cultures and traditions \citep{mckay2002}.

While significant advances have been made in the area of emotion detection and mood classification based on music and lyrics analysis, through large-scale machine learning operating on vast feature sets, sometimes spanning multiple domains, applied to relatively short musical selections \citep{kim2010}. Many times, these
approaches help in identifying the genre and mood but do not reveal much in terms of why a song is popular, or what features of the song made it catch the pulse of the audience. 

\cite{mihalcea2012} used LIWC and surface music components of all the phrases present in a small collection of songs as a dataset for identifying the emotions in that phrase. Each of the phrases was annotated for emotions. Using SVM classifier they obtained an accuracy of ~0.87 using just the language features. They observed that the language components gave a higher accuracy than music features in predicting emotions. The accuracy is higher as they are looking at emotions in a phrase, where the chance of having multiple emotions inside such a small text is very low.

When we look at a collection of popular songs, they belong to several emotional categories. It is clear from previous research that language is a strong indicator of emotions, but it is not clear if the language is an indicator of a song becoming a commercial success. 


We used the language features extracted from the lyrics to train an SVM classifier to identify the top and bottom category of songs. Below is the description of both approaches:

\begin{itemize}
  \item A machine learning approach: We extracted the language features, performed dimensionality reduction using principal component analysis (PCA) in-order to reduce the noise in the data. We trained and tested SVM classifier on the new features for identifying the songs that belonged to the top and bottom of the Billboard rankings.
    \end{itemize} 


\section{Data}
Billboard magazine \citep{billboard} is a world premier music publication since 1984. Billboard's music charts have evolved into the primary source of information on trends and innovation in music industry. With more than 10 Million users, its ranking is considered as a standard in the music industry. Billboard releases the weekly ranking of top $100$ songs in several categories such as rock, pop, hip-hop, etc. For this study, we used top $100$ hot-hits of every week from $2001 - 2010$. We collected the lyrics of the songs from \emph{www.lyrics.com}. Since the ratings of the songs are given
every week, there is a lot of repetition of the same song being in present in multiple weeks. For the simplifying the problem we selected the top rank of the song throughout the year as the rank of the song. 

\begin{figure}[h!]

\caption{Histogram of the best rank of songs from 2001-2010 of billboard top
100}

\centering

\includegraphics[width=0.8\textwidth,natwidth=610,natheight=642]{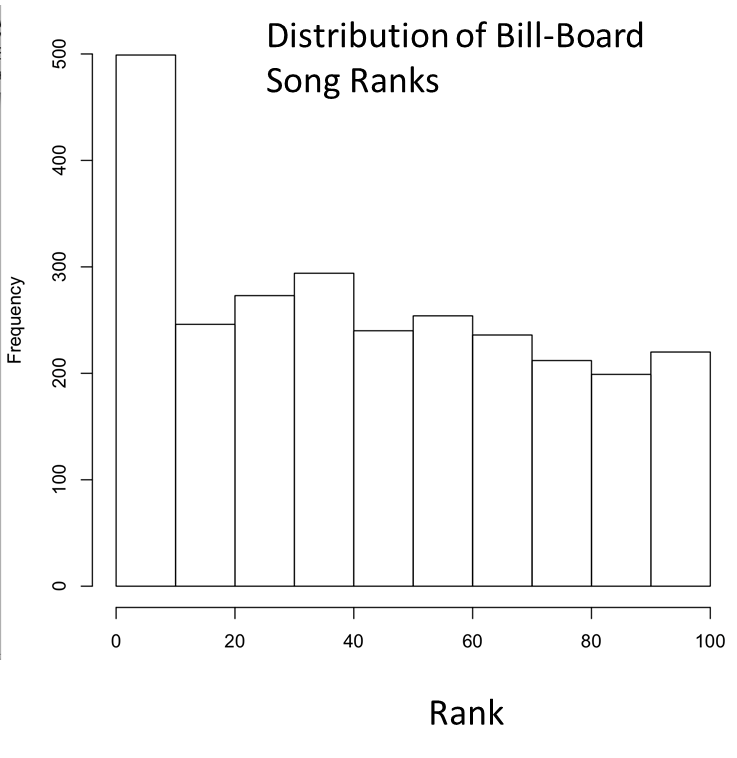}

\label{fig:hist}

\end{figure}

After cleaning the lyrics from hypertext annotations and punctuations, we had a total of $2683$ songs from $808$ artists. The histogram of the peak rank of the songs in the dataset is shown in Figure \ref{fig:hist}. For our analysis, we build a model to identify the songs that belonged to the top 30 and bottom 30 ranks. There are a total of $1622$ songs of which $991$ belonged to top 30, and the rest belonged to bottom 30 ranks.

\section{Features}

There are few analysis which conduct whole battery of linguistic algorithms that look at syntax, semantics, emotions, and affect contribution of words present in the lyrics. These algorithms can generally be classified into general structural (e.g., word count), syntactic(e.g., connectives) and semantic (e.g., word choice) dimensions of language, whereby some used a bag-of-word approach (e.g. LIWC), whereas others used a probability approach (MRC), whereas  yet  others 
relied  on  the computation  of  different factors (e.g., type-token ratio). There are eight computation linguistic
algorithms that are used to analyze the language features inside the lyrics of
the songs.

\begin{figure*}[ht]

\caption{Overview of computational linguistic algorithms used.
$^{1}$\cite{louwerse2001}, $^{2}$\cite{biber1991},
$^{3}$\cite{semin1988,semin1991}, $^{4}$\cite{johnson1989},
$^{5}$\cite{miller1998}, $^{6}$\cite{coltheart1981}, $^{7}$\cite{celex},
$^{8}$\cite{tausczik2010}}

\centering

\includegraphics[width=0.5\textwidth,natwidth=610,natheight=642]{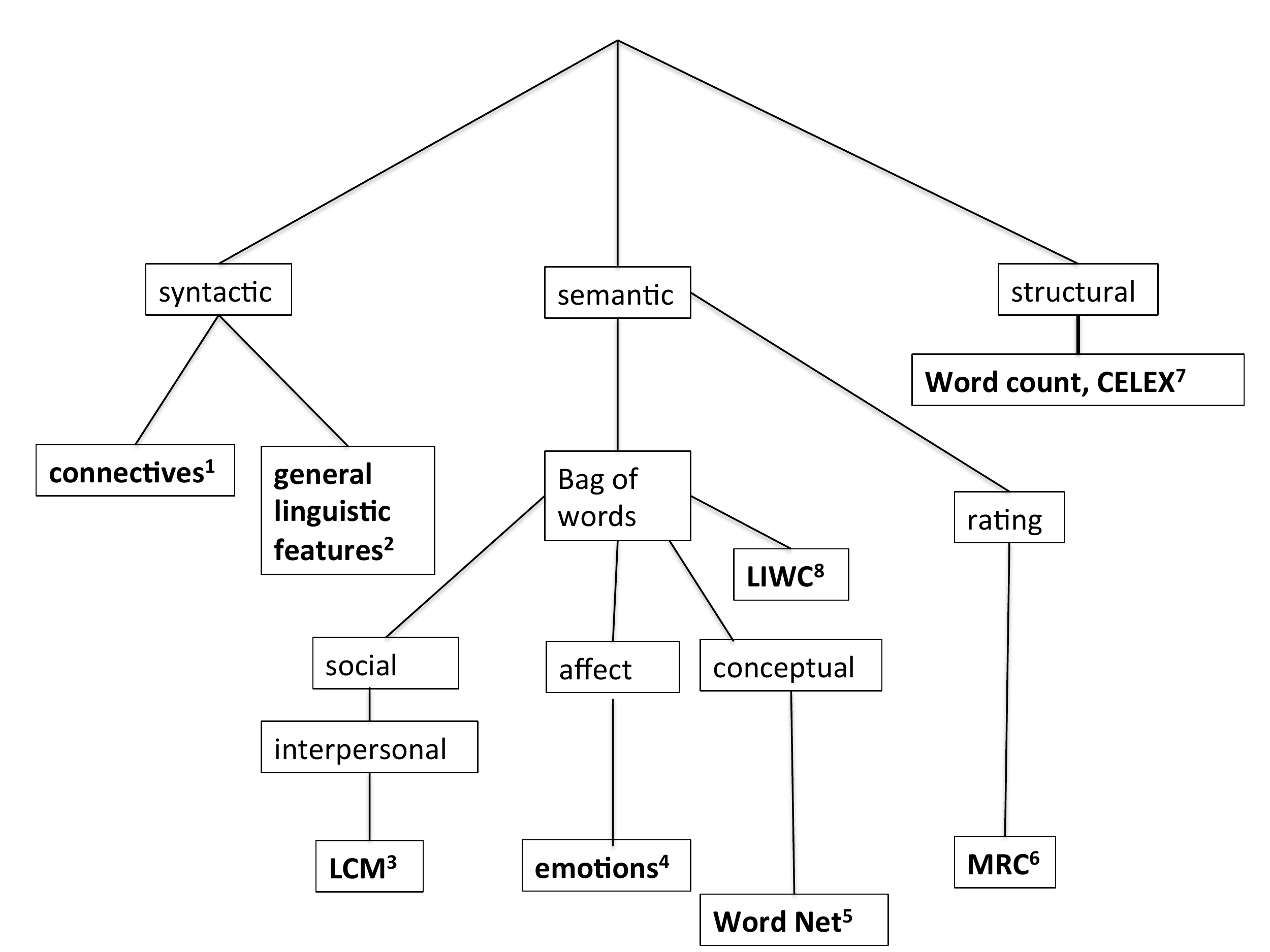}

\label{fig:computationalAlgorithms}

\end{figure*}

For general linguistic features, we used the frequency of $67$ linguistic features described by \citep{biber1991}. These features primarily operate at the word level (e.g., parts-of-speech) and can be categorized as tense and aspect markers, place and time adverbials, pronouns and proverbs, questions, nominal forms, passives, stative forms, subordination features, prepositional phrases, adjectives and adverbs, lexical specificity, lexical classes, modals, specialized verb classes, reduced forms and dis-preferred structures, and co-ordinations and negations \citep{luno2013}.

For semantic categories of the words, we used Wordnet \citep{miller1998}. Wordnet has $150000$ words in $44$ base types including $25$ primitive groups for nouns (e.g. time, location, person, etc.), $15$ for verbs (e.g. communication, cognition, etc.), $3$ groups of adjectives and $1$ group of adverbs. We also collected all the English words from Google unigrams \citep{brants2006} and binned them into one of the $44$ categories if one of their synonyms belonged to those categories. These words represent the categories such as communication nouns, social nouns, and many others.

The linguistic category model (LCM) gives insight into the interpersonal
language use. The model consists of a classification of interpersonal
(transitive) verbs that are used to describe actions or psychological states and adjectives that are employed to characterize persons. To capture the various emotions expressed by the statement, we have used the emotion words given by \citep{tausczik2010}, classified into two classes broadly basic emotions (anger, fear, disgust, happiness, etc.) and complex emotions (guilt, pity, tenderness, etc.).

The basic emotions indicate no cognitive load hence they are also called as raw emotions, whereas the complex emotions indicate cognitive load. Inter-clausal relationships were captured using parameterization, including positive additive, (also, moreover), negative additive (however, but), positive temporal (after, before), negative temporal (until), and causal (because, so) connectives. To get the frequencies of the words, we have used CELEX database \citep{celex}. The CELEX database consists of $17.9$ million words taken
from both spoken (news wire and telephonic conversations) and written (newspapers and books) corpora. Also, we used the MRC Psycholinguistic Database \citep{johnson1989}, to get linguistic measures such as familiarity, concreteness, and meaningfulness.

\section{Classification}

\begin{table}[ht]
\caption{Classification Results}
\centering
\begin{tabular}{c c c c}
\hline
Measures & Precision & Recall & Kappa \\ [0.5ex] 
\hline
SVM exp. ker &\textbf{0.76}&\textbf{0.76}&\textbf{0.51} \\
SVM ply. ker&0.68&0.68&0.36 \\
SVM lin. ker&0.53&0.53&0.05 \\

\hline
\end{tabular}
\label{table:classification1}
\end{table}

After the linguistic analysis, we approached the problem as a
classification problem. As discussed earlier, we extracted the language features from the lyrics using the computational linguistic algorithms shown in Figure ~\ref{fig:computationalAlgorithms}. We extracted 261 features from each of the 2616 songs. The goal is to build a classifier that predicts the top and bottom ranked songs of the Billboard. Since there are many features and very few songs, we removed the noise contributed by the features using principle component analysis (PCA). Features that explained 0.6 variance were selected, and this reduced the features to 39 from 261.

It is important to note that the major advantage of doing a PCA is noise
reduction, and also identifying the best features that capture the variance in the data. The disadvantage is that the variables loose their semantic meaning compared to the raw features. 

The classes of positive and negative samples i.e. the top 30 and bottom 30 songs were in the ratio of 1.5 to 1, and to balance the classes we performed synthetic minority over-sampling (SMOTE)
\citep{chawla2002}. The SMOTE creates new synthetic samples that are similar to the minority class by picking data points that are closer to the original sample. 

After balancing the classes, we performed classification using support vector machine (SVM) using a radial(exponential), polynomial and linear kernel functions. The classification is done using a 10-fold cross validation method.

SVM uses implicit mapping function defined by the kernel function, to map the input data into a very high dimensional feature space. Then it learns the plane of separation between the two classes of the high dimensional space. For the classification of top and bottom ranked songs we observe that the radial (exponential) function performs the best, with a precision ~0.76, recall ~0.76 and Cohen's Kappa ~-0.51. The kappa score indicates that the classifier did the classification with great confidence. 

We also attempted building a classifier using other classification algorithms such as Bayes, Naive-Bayes, and decision trees, but all of them performed poorly compared to the SVM.

\section{Discussion}
There are several studies \citep{mihalcea2012,su2013,laurier2008,kim2010} that
have looked into emotions in music based on language as well as few audio features. All the studies explicitly indicated that language features were more useful than surface music features in identifying the emotion present in the songs. 

Songs contain both music and lyrics. In this work, we have used only the lyrics as our data. Lyrics of the songs are available publicly when compared to the music. Since previous studies have shown the importance of language in music for identifying emotions, we extended the investigation for identifying the language features that help in differentiating the top and bottom rated songs on the billboard. To
the best of our knowledge this is a first study that uses computational
linguistic algorithms and machine learning models to predict whether a song belongs to top or bottom of the Billboard rankings.



We used the language features extracted using the language model to train SVM classifiers under different kernel functions to identify whether a song belongs to the top or bottom of the billboard chart. The radial kernel function gives a precision $0.76$with a kappa $0.51$ which indicates that the confidence in classification. 

Although audio features of the song play an important role, they are expensive and not publicly available for download. In this paper, we focused only on the language features and the results from both the studies indicate that we can robustly identify whether a song goes to top or bottom of Billboard charts based on the language features alone. Although the precision is only 0.76 (chance is 0.5), given that we are in a very dense space of top 100 songs from Billboard, where all the songs are best of the best when taking into consideration all the music albums
uploaded on to social media (youtube, facebook, twitter, etc.).

Overall the take-home message of this paper is that language features can be exploited by the machine learning algorithms to predict whether a song reaches the top or bottom of the Billboard rankings.

\section{Conclusion and Future Work}
The music industry is a vibrant business community, with many artists publishing their work in the form of albums, individual songs, and performances. There is a huge financial incentive for the businesses to identify the songs that are most likely to be a hit. 


can use machine learning models to train on several language features to predict whether a song belongs to the top 30 or bottom 30 of the Billboard ratings.

In future, we would like to expand our research question to predict whether the song reaches to the class of top 100 Billboard list or not.

\bibliographystyle{aaai}
\bibliography{music_bib}

\end{document}